\documentclass[letterpaper, 10 pt, conference]{ieeeconf}  

\IEEEoverridecommandlockouts                              

\usepackage{graphics} 
\usepackage{epsfig} 
\usepackage{mathptmx} 
\usepackage{times} 
\usepackage{amsmath} 
\usepackage{amssymb}  
\DeclareMathOperator*{\argmax}{argmax}
\usepackage{subcaption}
\usepackage{hyperref}
\title{\LARGE \bf
Vision Transformer for Learning Driving Policies in Complex Multi-Agent Environments
}

\author{Eshagh~Kargar,
        Ville~Kyrki
        \thanks{$^{1}$Eshagh Kargar and Ville Kyrki are with School of Electrical Engineering, Aalto University, Finland. {firstname.lastname}@aalto.fi}%
}

\begin{document}

\maketitle
\thispagestyle{empty}
\pagestyle{empty}

\begin{abstract}
Driving in a complex urban environment is a difficult task that requires a complex decision policy. In order to make informed decisions, one needs to gain an understanding of the long-range context and the importance of other vehicles.
In this work, we propose to use Vision Transformer (ViT) to learn a driving policy in urban settings with birds-eye-view (BEV) input images. 
The ViT network learns the global context of the scene more effectively than with earlier proposed Convolutional Neural Networks (ConvNets). Furthermore, ViT's attention mechanism helps to learn an attention map for the scene which allows the ego car to determine which surrounding cars are important to its next decision.
We demonstrate that a DQN agent with a ViT backbone outperforms baseline algorithms with ConvNet backbones pre-trained in various ways. 
In particular, the proposed method helps reinforcement learning algorithms to learn faster, with increased performance and less data than baselines.
\end{abstract}

\section{INTRODUCTION}

\begin{figure}[t!]
    \centering
    \includegraphics[width=\linewidth]{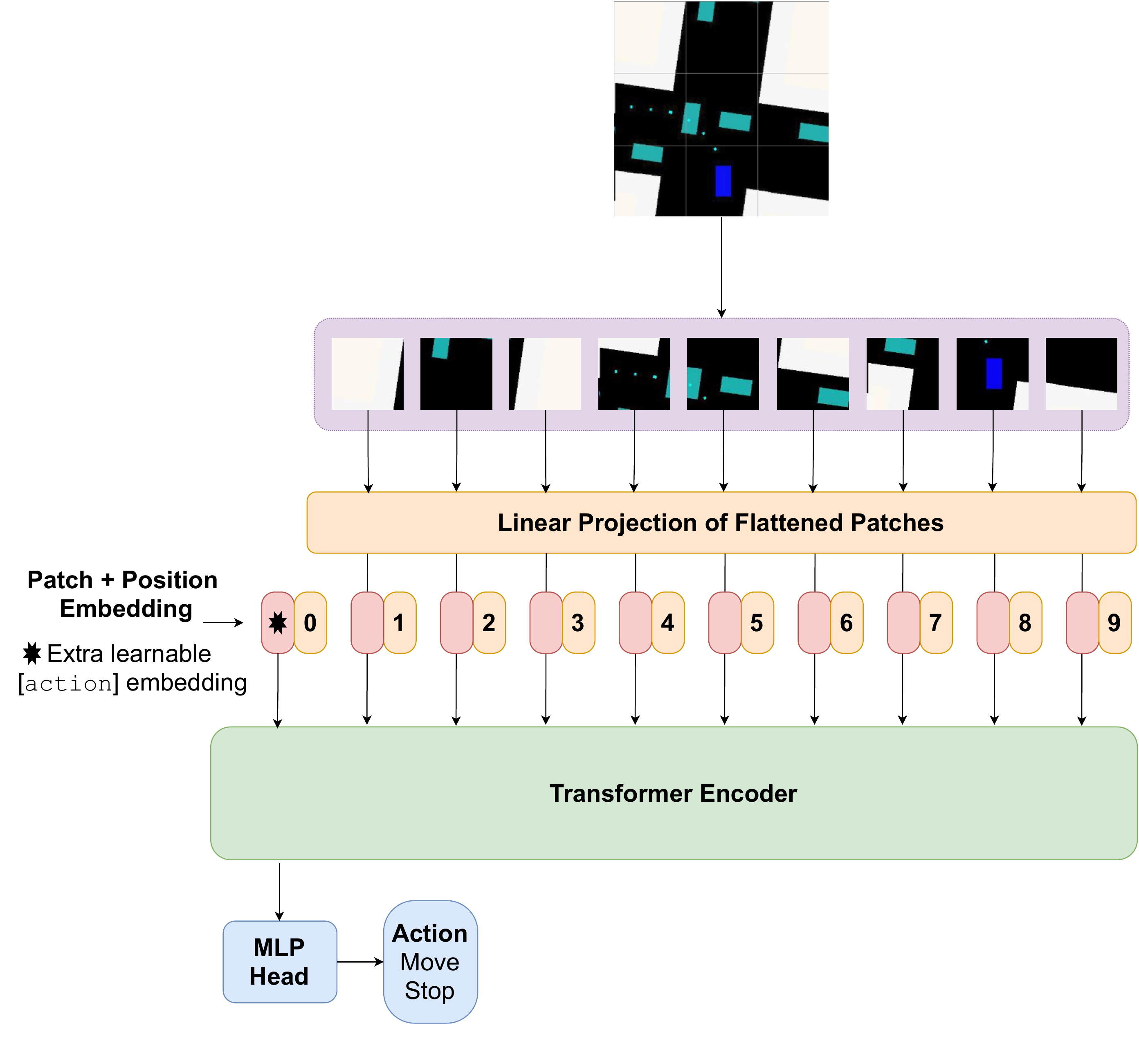}
    \caption{The proposed architecture with a Vision Transformer as backbone. We split the input top-down view image into fixed-size patches, linearly embed each of them, add position embeddings, and feed the resulting sequence of vectors to a standard Transformer encoder. In order to output an action, we add an extra learnable "action" token to the sequence. Also, we use the Deep Q-Network algorithm to train the model. The architecture for the transformer encoder is shown in Fig.~\ref{fig:transformer_encoder}.}
    \label{fig:vit}
\end{figure}

Driving in unstructured and dynamic urban environments is an arduous task. The decisions of drivers are influenced by many moving objects and the driver must be able to identify the importance and effect of each one on its behavior in order to act accordingly. Additionally, understanding map data and the global context of the environment around a car is essential for making decisions. As a result, autonomous driving in such complex environments remains a challenge for researchers and industry.

Reinforcement Learning (RL) is one of the emerging solutions to solve the decision-making problem in autonomous driving~\cite{kiran2021deep, kendall2019learning, saxena2020driving, ma2020reinforcement}. 
The use of RL in image-based environments has been widely explored~\cite{mnih2015human, kendall2019learning, laskin2020reinforcement, chen2019model}; however, the environments in which RL is used are usually single-agent environments, or if there are multiple agents, no multi-agent technique is used. In comparison, policy and interaction learning is more challenging in environments with multiple agents interacting with each other.

Convolutional Neural Networks (ConvNets) are used widely as they are good at capturing local context. However, they are not able to capture long-range context in the scene~\cite{gao2020vectornet} and lack an attention mechanism to handle multi-agent settings.

We propose to use Vision Transformers (ViTs) to learn a driving policy that can capture long-range context information and has an attention mechanism that allows the car to know where and how much to pay attention to. 
The attention is learned through interaction with other vehicles, without the need for heuristics or supervision. 
The driving policy uses birds-eye-view (BEV) images as input and outputs a high-level discrete action. 

The primary contributions of this work are: (a) using a Vision Transformer in Autonomous Driving with vehicles and map data rendered as a BEV image to learn a global context of the scene and where to pay attention to, (b) an "action" token which allows to use Vision Transformers for RL, and (c) evaluation in an urban driving scenario with multiple interacting vehicles, showing better performance, faster convergence, lower crash percentage, and meaningful attention map for multiple cars in the scene in situations such as four-way and T-intersections.


\section{Related Work}\label{related_work}

Learning a decision-making policy using reinforcement learning is a popular approach in autonomous driving ~\cite{chen2019model, saxena2020driving, ma2020reinforcement, kendall2019learning}. Although RL is typically data-hungry, Kendall et al.~\cite{kendall2019learning} recently demonstrated the learning of a lane following task using real-world data gathered in a single day. Nevertheless, the complexity and multi-agent interaction of complex traffic environments require a method that takes into account both agent-agent and agent-environment interactions.

Autonomous driving systems also rely on HDMap data to make decisions. 
Map data can be rendered on a BEV image and encoded using ConvNets~\cite{chen2019model,bansal2018chauffeurnet}. 
While ConvNets are very good at capturing local patterns, the global context can not be captured very well.
Graph neural networks were proposed by Jiyang et al.~\cite{gao2020vectornet} to capture better global context and encode map and agent information. 

In addition to the global context, which is crucial for decision-making, vehicles interacting with the ego car are also crucial. However, not all cars and not all parts of the scene are equally important. 
Therefore, an attention mechanism can be useful in paying more attention to important vehicles and critical parts of the map in the decision-making problem.
Attention mechanism can be used with ConvNets to improve explainability and interpretability of end-to-end deep neural networks~\cite{cultrera2020explaining, kim2017interpretable}. 
A multi-task attention-aware network with a ConvNet backbone was proposed by Ishihara et al.\cite{ishihara2021multi} to learn a driving policy via conditional imitation learning. 
Chen et al.~\cite{chen2019attention} designed a hierarchical Deep Reinforcement Learning algorithm and used an attention mechanism with ConvNets to learn faster, safer, and smoother lane change behaviors. 

In addition to using of attention mechanisms with ConvNets, recently, Transformer architecture has become the de-facto standard for natural language processing tasks~\cite{vaswani2017attention} and has been used in computer vision applications~\cite{dosovitskiy2020image}. However, its application in RL remain limited and is not studied. Leurent et. al~\cite{leurent2019social} used Transformers in an RL task for autonomous driving but with state-vector inputs. Transformers are also used in some works for the motion prediction task~\cite{liu2021multimodal, giuliari2021transformer}.

In this work, we propose to use Vision Transformers as backbone for RL methods in multi-agent environments which could be used to handle both (1) the multi-agent nature of the environment via its attention mechanism, (2) and encoding the map and vehicles' information simultaneously, as well as capturing global context.

\section{Background}\label{background}

In this section, we will outline the required theoretical concepts behind the proposed method, including Reinforcement learning, Deep Q-Networks, and Transformers. 

\subsection{Reinforcement Learning}
Reinforcement learning aims to find an optimal control policy $\pi:S\rightarrow A$ from states to actions that maximizes total expected future rewards
\begin{equation}\label{rl_reward}
     R(\pi) = E_{\pi} \Bigg[ \sum_{t} \gamma^t r(s_t, a_t) \Bigg] 
\end{equation}
where \(s_t\), \(a_t\), \(r\), and \(\gamma\) are state at time \(t\), action at time \(t\), reward, and discount factor, respectively.

Value-function based RL solves the RL problem by determining an optimal value function
\(Q:S,A\rightarrow \mathbb{R}\) that describes the expected cumulative rewards when starting from a particular state and choosing a particular action
\begin{equation}
     Q^*(s, a) = \max_{\pi} E\Bigg[ \sum_{t=0} \gamma^t r(s_t, a_t) |s_0=s, a_0=a \Bigg] .
\end{equation}
Knowing the optimal value function, an optimal  policy \(\pi^*\) can then be determined as \(\pi^*(s) = \argmax\limits_{a} Q^*(s,a) \).

\subsection{Deep Q-Network}
It is usually impossible to determine the value function of continuous state spaces with unknown dynamics. 
Deep Q-Networks (DQNs)~\cite{mnih2015human} are a successful RL algorithm that use a deep neural network \(Q(s, a; \psi) \) to approximate the value function where \(\psi\) are the parameters of the neural network. 
DQN also stores past experiences in a replay buffer $D=\{(s,a,r,s') \}$ to accelerate and stabilize learning. 
To further stabilize the learning, DQN defines a target Q-network with parameters \( \psi' \) which are updated only every \(\tau\) steps to the current \(\psi\).  
To optimize \(\psi\), the Q-learning loss
\begin{equation}
     L_{DQN}(\psi) = E_U(D)\Bigg[ \left(r+\gamma \max_{a'}Q(s',a';\psi') - Q(s,a,\psi)\right)^2 \Bigg]
\end{equation}
is minimized for a uniform sample of transitions sampled from $D$.

\subsection{Transformers and the Vision Transformer}
Transformer architectures with attention mechanism are commonly used in sequence modeling tasks such as natural language processing (NLP). 
With the attention mechanism, dependencies can be modeled without regard for their distance in the input or output sequence.
Moreover, Transformers' parallelization makes them more efficient than ConvNets and Recurrent Neural Networks, and they can train significantly faster. Additionally, they are very adept at identifying global dependencies between inputs and outputs~\cite{vaswani2017attention}.

\begin{figure}[t!]
    \centering
    \includegraphics[width=\linewidth]{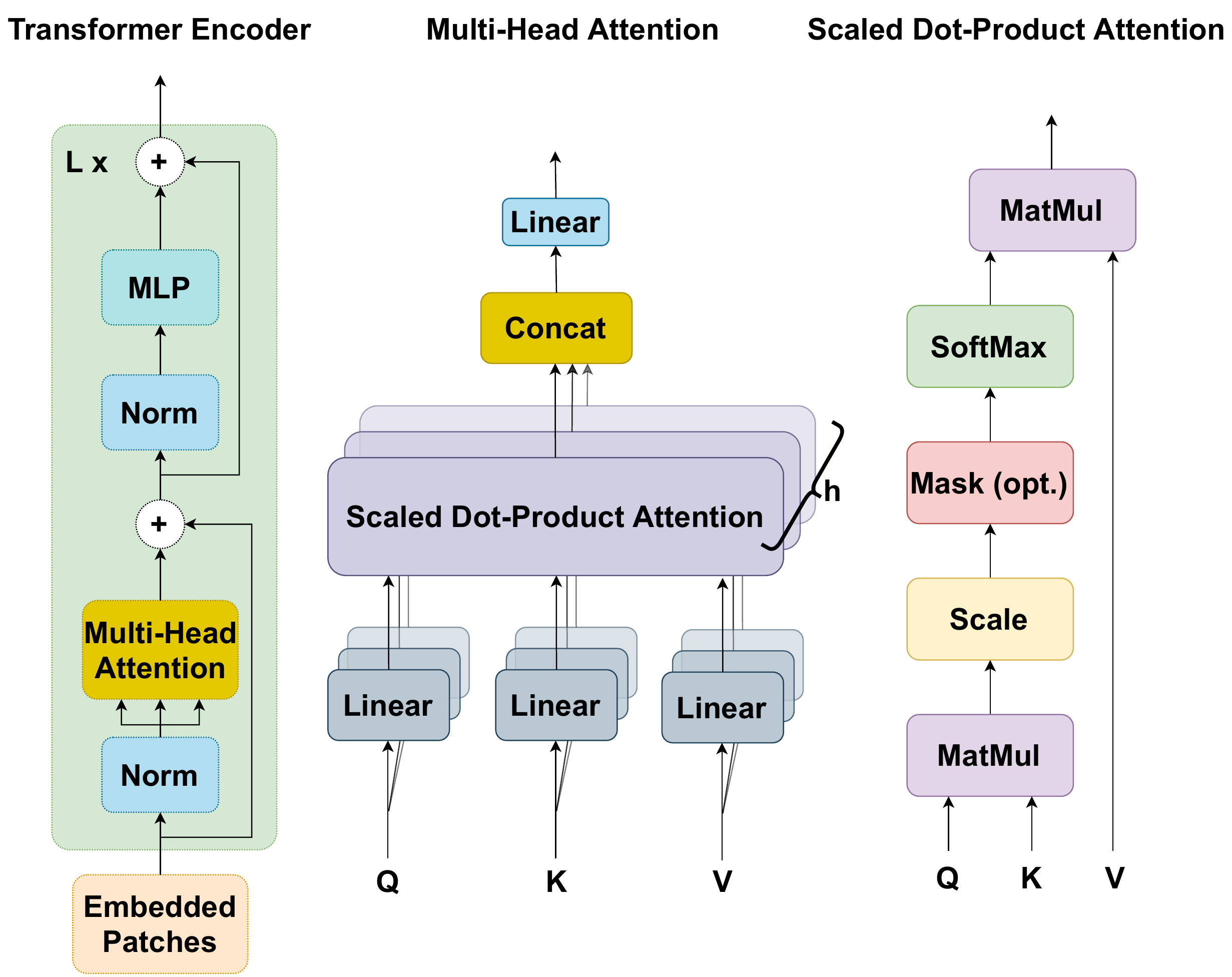}
    \caption{(left) Transformer Encoder architecture with L stacked layers. (middle) Multi-Head Attention with several attention layers running in parallel. (right) Scaled Dot-Product Attention. The architecture is inspired by \cite{vaswani2017attention}.}
    \label{fig:transformer_encoder}
\end{figure}

Transformers have a structure of encoders and decoders. In this work we only use the encoder part, illustrated in Fig.~\ref{fig:transformer_encoder}.
The transformer encoder consists of $L$ stacked identical layers, each with a multi-head attention layer and a fully connected layer. 
Both sub-layers have residual connection and layer-normalization. 

In computer vision, a variant called Vision Transformers~\cite{dosovitskiy2020image} (ViT) has been used extensively. 
In order to feed a $2D$ image
into the Transformer model, ViT~\cite{dosovitskiy2020image} converts the input image into non-overlapping patches.
The $2D$ patches are then flattened and mapped into a vector using a trainable linear projection. The patch embeddings are then concatenated with learnable 1D position embeddings, which preserve positional information, and fed into the Transformer Encoder. 

\section{Method}\label{method}

In this section, we first explain the problem setting and review the proposed idea to solve it. 
We then proceed with describing a detailed description of the proposed method including the Vision Transformer equations and the RL algorithm on top of that. 
Finally, we present the input representation. 

\subsection{Problem Setting and Solution Overview}
The map information and other vehicles' information are rendered on a BEV RGB image $x \in R^{H \times W \times C}$ which serves as the input. 
The input is then processed by a Vision Transformer which generates a high level discrete action as output. 
The Vision Transformer is trained using the DQN algorithm. 
The structure is illustrated in Fig.~\ref{fig:vit}.



\subsection{Vision Transformer}

In order to feed the input image $x \in R^{H \times W \times C}$ into the Vision Transformer, we reshape the BEV image into flattened patches $x_p \in R^{N \times (P^2C)}$, where $(H, W)$ is the input image resolution, $C$ is the number of channels, $(P, P)$ is the size of each patch, and $N = HW/P^2$ is the number of patches and the input sequence length. 

In order to use Vision Transformer in RL, we propose to prepend a learnable "action" token ($\textbf{z}_0^0 = x_{action}$) to the input sequence of embedded patches. The idea is inspired by the original Vision Transformer work which used a "class" token~\cite{dosovitskiy2020image} for classification tasks. 

The Transformer encoder module is shown in Fig.~\ref{fig:transformer_encoder} and its mathematical formulation can be written
\begin{equation}
\begin{split}
    \textbf{z}_0 = & [\textbf{x}_{action}; \textbf{x}_p^1 \textbf{E}; \dots ; \textbf{x}_p^N \textbf{E}] + \textbf{E}_{pos}, \\
    & \quad \textbf{E} \in R^{(P^2.C) \times D}, \quad \textbf{E}_{pos} \in R^{N+1} \times D 
\end{split}
\end{equation}
\begin{equation}
    \textbf{z}'_l = MSA(LN(\textbf{z}_{l-1})) + \textbf{z}_{l-1}
\end{equation}
\begin{equation}
    \textbf{z}_l = MLP(LN(\textbf{z}'_l)) + \textbf{z}'_l
\end{equation}
where $LN$ is Layernorm layer~\cite{ba2016layer}, $MSA$ is a Multi-head Self Attention block, and $MLP$ block contains two linear layers with a GELU non-linearity.

The $MSA$ block is 
\begin{equation}
    [\textbf{Q}, \textbf{K}, \textbf{V}] = \textbf{z} \textbf{U}_{QKV}, \quad \textbf{U}_{QKV} \in R^{D \times 3D_h} 
\end{equation}
\begin{equation}
    A = softmax(\textbf{Q} \textbf{K}^T / \sqrt{D_h} ), \quad A \in R^{N \times N}
\end{equation}
\begin{equation}
    SA(\textbf{z}) = A\textbf{v}
\end{equation}
\begin{equation}
    MSA(\textbf{z}) = [SA_1(\textbf{z}); SA_2(\textbf{z}); \dots ;SA_k(\textbf{z})]\textbf{U}_{msa}, \quad \textbf{U}_{msa} \in R^{k.D_h \times D} 
\end{equation}

The first output vector of the Transformer Encoder ($\textbf{z}_L^0$) corresponding to the learnable "action" token is used to output the action:
\begin{equation}
    \textbf{a} = MLP(\textbf{z}_L^0).
\end{equation}

The whole network can be trained using any RL algorithm. Here we have discrete actions and used DQN to train the network weights. The proposed method is easily extendable to continuous action spaces and other RL algorithms. The whole proposed method is depicted in Fig.~\ref{fig:vit}.

\subsection{Input Bird-Eye View Image}
As mentioned before, one way to consider map information in the decision making process is to render them on a BEV image. It is also possible to render all the agents such as vehicles, pedestrians, and bicycles on the same image.

The rendered data on the BEV input image are as follows:

\begin{enumerate}
    \item \textbf{HDMap}: The HD-Map information includes drivable areas, road boundaries, and curbs. It is also possible to render more information such as lane lines and center of lane lines. Due to our intention to do longitudinal control only, this information is not used here. However, it is extremely useful for lateral control and steering angle generation.

    \item \textbf{Route}: The planned route to the destination is assumed to come from an external planning system such as a standard vehicle navigator.

    \item \textbf{Pose of other vehicles}: The current position, orientation, and size of other cars are rendered on the BEV image. This information can come from the perception module with object detection algorithms.

    \item \textbf{Pose of the ego vehicle}: The current position, orientation, and size of the ego vehicle are also rendered on the BEV image. This information can come from the Localization module.
\end{enumerate}

The generated BEV image will be used as input to the ViT network as shown in Fig.~\ref{fig:vit}.

\section{Experiments}\label{experiments}

In this section, we first introduce the simulator we used to do our experiments. Then we study the following questions:
\begin{enumerate}
    \item Do ViT-based DQNs improve the performance compared to ConvNet-based DQNs? \label{q1_baselines}
    \item How the proposed ViT-based DQN method performs compared to state-of-the-art RL algorithms? \label{q2_baselines2}
    \item How ViT backbone helps to pay attention to different parts of the scene and other cars in the environment and how the ego car reacts to them? \label{q3_attnmap}
    \item What is the effect of patch size in the Vision Transformer? \label{q4_psize}
\end{enumerate}

\subsection{Simulation}
\begin{figure}[h]
    \centering
    \includegraphics[width=0.7\linewidth]{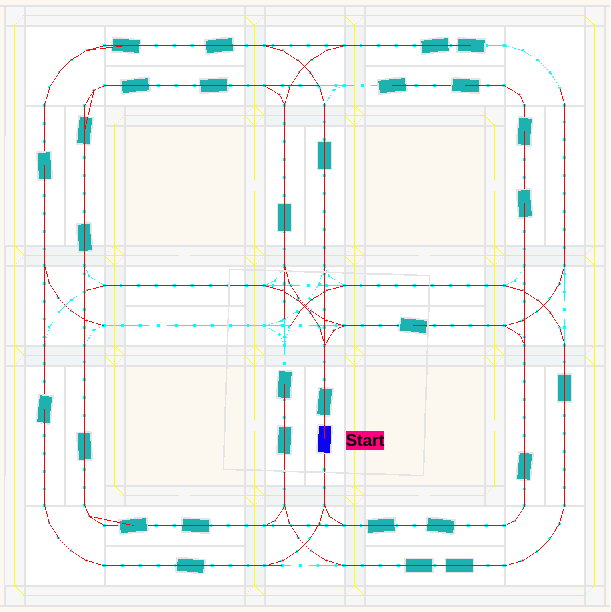}
    \caption{Fluids simulation environment.}
    \label{fig:fluids}
\end{figure}

In order to test our proposed method in a multi-agent setting, we use Fluids~\cite{zhao2018fluids} which is a self-driving car simulator. Fig.~\ref{fig:fluids} shows the simulation environment.

Fluids simulator supports intersection, left and right turn scenarios with and without traffic light. It also has some capabilities to add pedestrians and enable traffic lights for cars and pedestrians. In this work, in order to study the interaction between cars in a complex scenario, we spawned 30 cars randomly in the city. We also spawned the ego car, the blue one, before an unsignalized intersection. The destination for the ego car is a random point in the city. Also, we disabled all traffic lights and did not spawn any pedestrian in the city. 
The state space is a $80 \times 80$ top-down view image which is shown in Fig.~\ref{fig:vit} and the actions are just two commands, move with a constant velocity or stop, to control the longitudinal movement. For lateral control we used a PID controller to generate steering angle. 

The reward function is a sum of two terms: collision ($r_c$) and travelled distance ($r_d$),
\begin{equation}\label{eq:reward}
    r = r_c + r_d
\end{equation}
where $r_c$ is -100 if a collision happens and 0 otherwise, and $r_d$ is the travelled distance from starting point in meters. The first term encourages the car to avoid collision, and the second term pushes it to go forward and prevents it from stopping.

\subsection{Implementation Details}
Our proposed method has two parts: a ViT backbone and an MLP head for action outputs. For the ViT backbone, we used a ViT-small version used in~\cite{caron2021emerging} with 12 Transformer Encoder blocks and 6 heads in its multi-head attention layers. The dimension of projected layers and the output of Transformer Encoder layers is 384. 
The MLP head consists of two linear layers of size 64 and 2 for two actions with GELU non-linearity.

\subsection{Ablation Study}
In order to evaluate the performance of the proposed ideas and answer questions \ref{q1_baselines} and \ref{q4_psize}, we used DQN algorithm with various backbones to encode the input. 
As the input is an image and needs to be encoded, we considered several ConvNets, pre-trained in various ways, as the backbone for the DQN agent and fine-tuned them on our task. We selected ResNet-50 because of its similarity with ViT-small along several axes: number of parameters, throughput, and supervised performance on ImageNet~\cite{caron2021emerging}. We also considered ResNet-18 which is a smaller version of ResNet-50 as a baseline to see the effect of model size.

We show two versions of our proposed method in the comparison:
\begin{itemize}
    \item \textbf{ViT-scratch-p2,4,5,8,10}: ViT-small with $pathc\_size = 2, 4, 5, 8, 10$ and without pre-training phase. This version is presented to evaluate the effect of the pre-training stage and patch size in the proposed method's performance. 
    \item \textbf{ViT-DINO}: ViT-small with $patch\_size = 8$ pre-trained on ImageNet dataset using a self-supervised algorithm called DINO~\cite{caron2021emerging}. However $patch\_size = 4$ outperforms other versions, there was no available pre-trained model for that and we used $patch\_size = 8$.
\end{itemize}
These are compared against the following ablations with ConvNet backbones and DQN algorithm are as follows:
\begin{itemize}
    \item \textbf{ResNet18-ImageNet}: ResNet-18 pre-trained on ImageNet in a supervised learning classification task.
    \item \textbf{ResNet50-ImageNet}: ResNet-50 pre-trained on ImageNet in a supervised learning classification task.
    \item \textbf{ResNet18-VAE}: ResNet-18 pre-trained on ImageNet in a supervised learning classification task and then fine-tuned using a VAE (used as the encoder) on a collected dataset of size 50k from Fluids environment to reconstruct the input scene~\cite{chen2019model}.
    \item \textbf{ResNet50-VAE}: ResNet-50 pre-trained on ImageNet and then fine-tuned using a VAE (used as the encoder) on a collected dataset of size 50k from Fluids environment to reconstruct the input scene~\cite{chen2019model}.
    \item \textbf{ResNet50-DINO}: ResNet-50 pre-trained on ImageNet dataset using a self-supervised algorithm called DINO~\cite{caron2021emerging}.
\end{itemize}
The whole network, the backbone and the MLP head, is then fine-tuned using DQN algorithm with the same hyperparameters for all cases.

To evaluate the performance of the proposed method with ablations, we report two types of results:

\paragraph{Evaluation reward during the training phase} In each training epoch, the trained model is evaluated for two episodes and the average achieved reward is reported. 
\paragraph{Crash Percentage} After the training is finished, the trained model for each method is used to be run in the simulation for 200 episodes and number of crashes are reported. In order to have a fair comparison, the same set of random seeds are used.

\begin{figure}
    \centering
    \includegraphics[width=\linewidth]{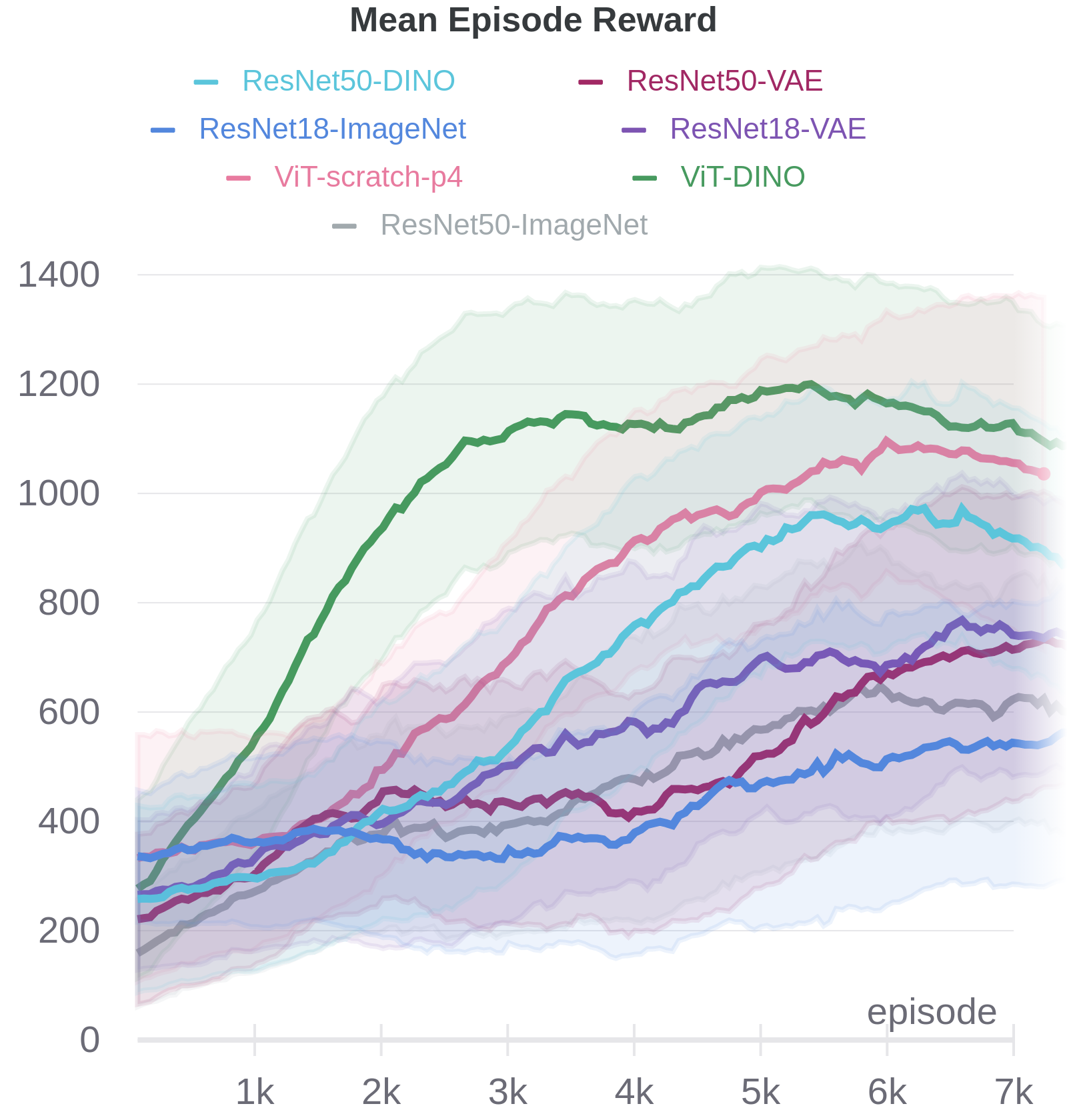}
    \caption{Comparison of our methods against DQN baselines in an unsignalized 4-way intersection.}
    \label{fig:baselines}
\end{figure}

Fig.~\ref{fig:baselines} shows the evaluation reward during the training phase. We just reported the mean episode reward for ViT-scratch-p4, because it performs better than other patch sizes and also to keep the plots clean.
ViT-DINO has the best mean evaluation reward of around 1200. The next best performance is for ViT-scratch-p4 which has a maximum mean evaluation reward of 1100. ResNet-DINO follows these two ViT-based models.
As can be seen in the plots, ViT-DINO reaches the performance of ViT-scratch-p4 and ResNet-DINO with less than half and one-third of their training data, respectively. This shows the effect of pre-training and then fine-tuning on a down-stream task with much less data. It also indicates the impact of the ViT backbone even without a pre-training phase. 
The performance can improve if we pre-train the model using DINO on a large driving dataset, which is available to most of self-driving car companies.

Other methods with ConvNet backbones are after these ViT-based models. ResNet50-DINO outperforms ResNet50-ImageNet which shows the effect of the pre-training method and better learned representation. 
We can also see a small improvement of ResNet18-VAE and ResNet50-VAE over ResNet18-ImageNet and ResNet50-ImageNet that demonstrates training of the ResNet models as encoder in a VAE can lead to a better representation and improve the performance.
It can also be seen that ViT-scratch-p4 can reach the best performance of ConvNet methods with almost half of their training data which shows the effect of ViT backbone even without pre-training phase.

\begin{figure}[t!]
    \centering
    \includegraphics[width=\linewidth]{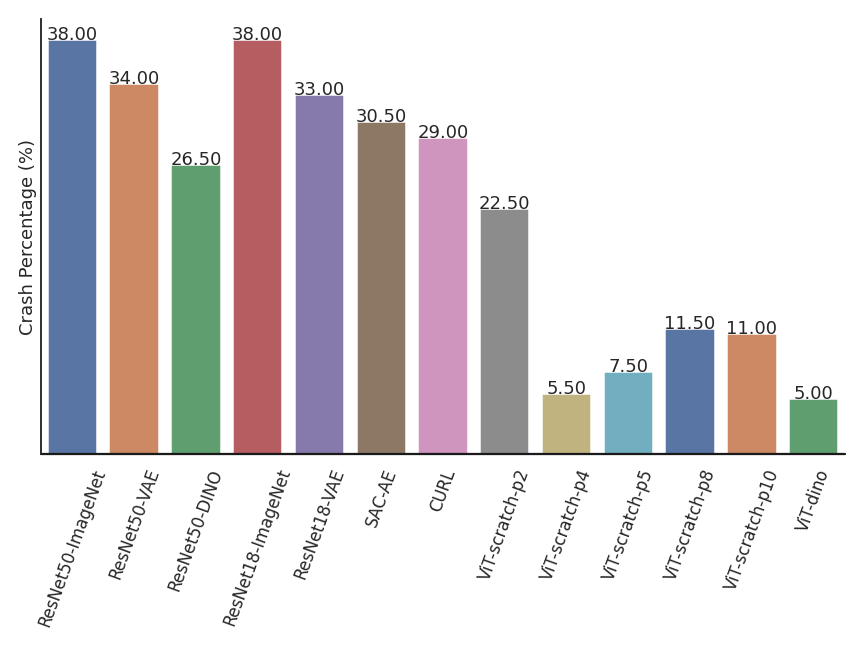}
    \caption{Crash Percentage in 200 runs.}
    \label{fig:crash}
\end{figure}

In addition, as shown in Fig.~\ref{fig:crash}, ViT-DINO has the best performance compared to other methods with the crash percentage of $5\%$. 
For ViT backbones trained from scratch, as we decrease the patch size, we see a small improvement in crash percentage, but for smaller patch sizes than 4, the crash percentage increases again. 
The next best performance after ViT-DINO is for ViT-scratch-p4 with the crash percentage of $5.5\%$. 
In general, the performance of all ViTs with different patch sizes are better than ConvNet methods. However, ResNet50-DINO has a lower crash percentage compared to other ConvNet-based methods which indicates the effect of the pre-training method.

The results represent the performance improvement caused by using ViT compared to ConvNet backbones, whether it is pre-trained or not. 

Videos of example driving using the learned model, DQN agent with ViT-DINO backbone, can be found at \href{https://youtu.be/-9TBLZEeCDs}{\textit{https://youtu.be/-9TBLZEeCDs}}. 

\subsection{Comparison to State of the Art}
\begin{figure}
    \centering
    \includegraphics[width=\linewidth]{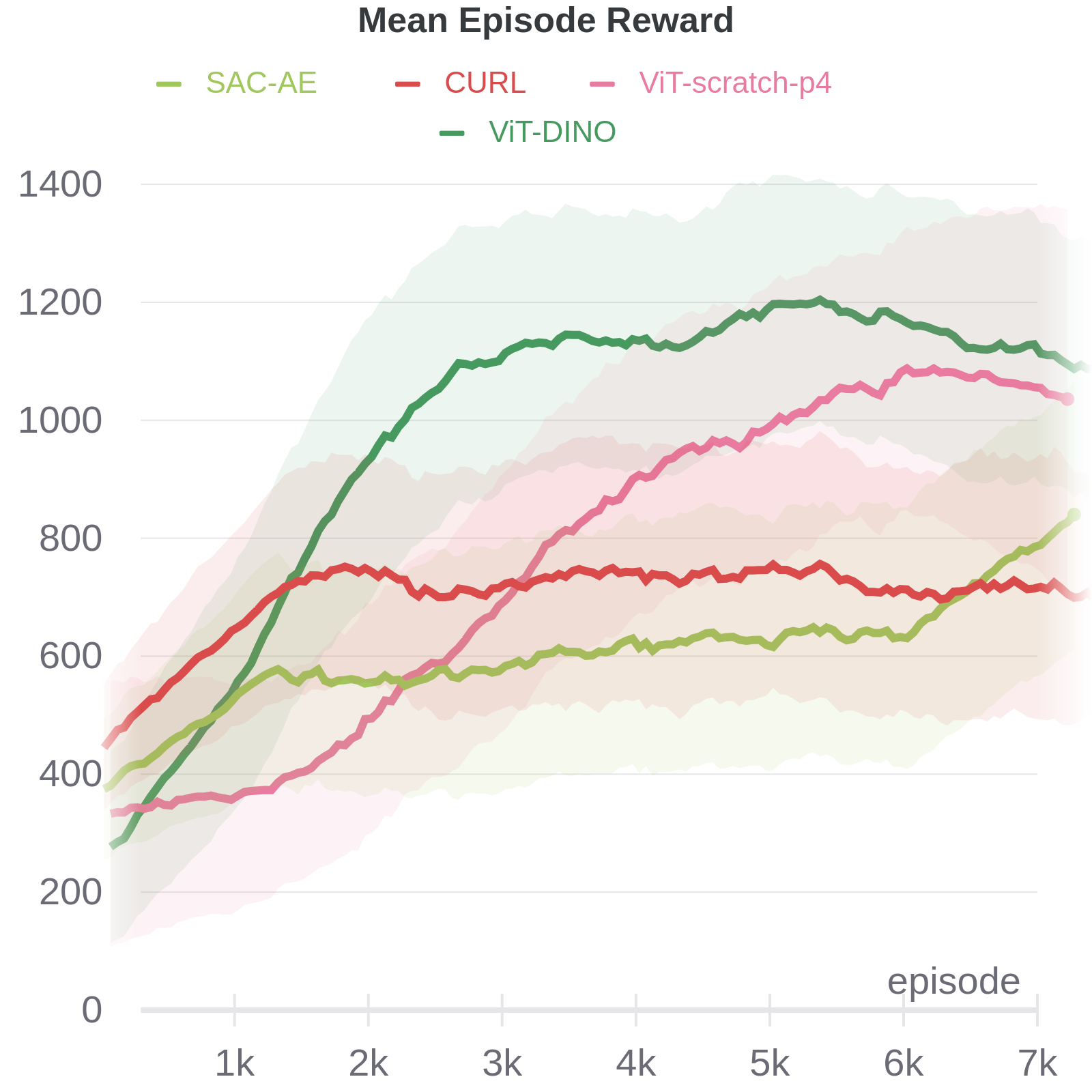}
    \caption{Comparison of our methods against state-of-the-art baselines in an unsignalized 4-way intersection.}
    \label{fig:baselines2}
\end{figure}

In order to answer question \ref{q2_baselines2}, we compared the proposed approach to two other state-of-the-art RL algorithms capable of handling image inputs, SAC-AE~\cite{yarats2019improving} and CURL~\cite{srinivas2020curl}. These two methods use ConvNets in their architecture and output a continuous action. The Fluids simulator's environment was modified to receive continuous velocity commands in order to support these methods.
We used the hyperparameters from the main papers for these two baselines.
Fig.~\ref{fig:baselines2} depicts the superior performance of the proposed method, trained from scratch or pre-trained with DINO, compared to SAC-AE and CURL. 

Furthermore, the crash percentage reported in Fig.~\ref{fig:crash} indicates that the proposed method performs significantly safer.
This shows the effect of the attention mechanism and the improved learning of the global context in ViTs. 
CURL, SAC-AE, and any other RL algorithm can easily use the ViT instead of ConvNet as their backbone.

\subsection{Visualize the Learned Attention Map in ViT}
\begin{figure}
     \centering
     \begin{subfigure}[b]{0.49\columnwidth}
         \centering
         \includegraphics[width=0.9\columnwidth]{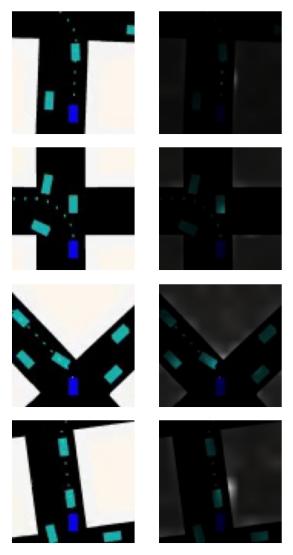}
         \caption{}
         \label{fig:attnmap_viz_a}
     \end{subfigure}
     \hfill
     \begin{subfigure}[b]{0.49\columnwidth}
         \centering
         \includegraphics[width=0.9\columnwidth]{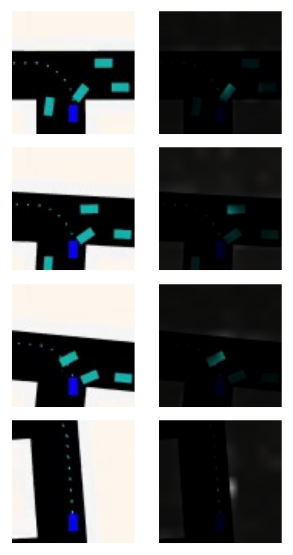}
         \caption{}
         \label{fig:attnmap_viz_b}
     \end{subfigure}
        \caption{Attention map visualization for two scenarios. (a) unprotected left-turn in a four-way intersection, (b) unprotected left-turn in a T-intersection. In each scenario, the left column shows the raw BEV image. For better visual understanding, the right column depicts the learned attention map as it is rendered on the BEV image. High attention is indicated by white areas in the images.}
        \label{fig:attnmap_viz}
\end{figure}

In order to understand the attention map in ViT models and answer question \ref{q3_attnmap}, we present visualizations of it in two unprotected left-turns, which are among the most challenging in autonomous driving. Fig.~\ref{fig:attnmap_viz} illustrates these two examples. ViT-DINO, the best model from the previous section, was used for this visualization.

In the 4-way intersection scenario, Fig.~\ref{fig:attnmap_viz_a}, the ego car does not pay attention to any other vehicle in the first image. In the second image, as it approaches the intersection, it begins to pay attention to the car in front of it and a little bit to the car coming from the top. The third image shows that the ego car concentrates more on what is in front of it and its left side, which are more critical than other cars in that state. At the end, when the ego car passes the intersection, it just pays attention to its lead vehicle and follows it. Attention values in each image are relative and should be interpreted independently.

In the T-intersection scenario shown in Fig.~\ref{fig:attnmap_viz_b}, the ego car pays attention to its lead vehicle first. Then, in the second and third images, it focuses on the upcoming car, which wants to take a left turn and is more critical than other cars for the left-turn decision of the ego car. The ego car waits for the upcoming car to pass, then makes the left turn and continues on its way in the last image. 

\section{CONCLUSIONS}\label{conclusion}
 
Reinforcement learning presents an important avenue to handle the complexity of situations in autonomous driving with multiple interacting cars and map information that need to be encoded and take into account in decision-making, but its applicability is significantly hindered by the information encoding method. 
In this paper, we proposed a framework to use Vision Transformers in RL settings with multiple interacting agents. The map and vehicles' information can be rendered on a BEV images. Vision Transformers can encode this information while learning the global context and an attention map to identify which car in the environment is more important than others to the ego car. 
Experimental comparison against ablations and baselines showed that the use of Vision Transformer improves policy learning in several ways: the policy quality is better, the learning converges faster, and the learned policy is safer and has a lower crash percentage.
The proposed approach was tested using DQN algorithm and can be easily extended to other reinforcement learning methods for both discrete and continuous action spaces.

Using the proposed approach in real traffic environments benefits from a pre-training step on real world datasets. 
The pre-training improves the quality of the policy and reduces the need for real world interaction in the reinforcement learning phase. 
Real world also has many constraints such as traffic rules that have to be taken into account. 
This may pose a challenge for methods based on learning, also because learning methods may lack interpretability, even if that can be alleviated partly by techniques such as the attention mechanism proposed in this paper. 
Thus, integration of logic and knowledge with learning based systems remains the greatest open issue in building robust and generalizable decision making for autonomous cars.

\addtolength{\textheight}{-10cm}   



\section*{ACKNOWLEDGMENT}
The authors wish to acknowledge CSC – IT Center for Science, Finland, for generous computational resources.
We also acknowledge the computational resources provided by the Aalto Science-IT project.


\bibliographystyle{IEEEtran}
\bibliography{ref}


\end{document}